%% file: acl_latex.tex
\documentclass[11pt]{article}

\usepackage[final]{acl}

\usepackage{times}
\usepackage{latexsym}

\usepackage[T1]{fontenc}

\usepackage[utf8]{inputenc}

\usepackage{microtype}

\usepackage{inconsolata}

\usepackage{graphicx}
\usepackage{float}

\usepackage{booktabs}
\usepackage{multirow}
\usepackage[table]{xcolor}
\usepackage{adjustbox}
\usepackage{amsmath,amssymb}
\usepackage{algorithm}
\usepackage{algpseudocode}

\definecolor{modelbg}{RGB}{242,244,247}   
\definecolor{oursbg}{RGB}{226,237,252}    

\usepackage[most]{tcolorbox}
\tcbuselibrary{listings,breakable,skins}
\usepackage{listings}
\usepackage{xcolor}
\usepackage{enumitem}

\definecolor{promptHeader}{HTML}{4A4A4A}
\definecolor{promptBorder}{HTML}{B7C7E5}
\definecolor{promptPanel}{HTML}{EEF3FB}
\definecolor{promptPanelDark}{HTML}{DCE6F6}
\definecolor{promptCode}{HTML}{F7F9FC}
\definecolor{promptBlue}{HTML}{3E67B1}
\definecolor{promptPurple}{HTML}{9867C5}

\newcommand{\promptslot}[1]{\textcolor{promptBlue}{\texttt{\{#1\}}}}
\newcommand{\prompttag}[1]{\textcolor{promptPurple}{\texttt{#1}}}

\tcbset{
    promptouter/.style={
        enhanced,
        colback=white,
        colframe=promptBorder,
        colbacktitle=promptHeader,
        coltitle=white,
        fonttitle=\bfseries\large,
        boxrule=0.8pt,
        arc=4mm,
        left=2mm,
        right=2mm,
        top=2mm,
        bottom=2mm,
        toptitle=2mm,
        bottomtitle=2mm
    },
    promptpanel/.style={
        enhanced,
        colback=promptPanel,
        colframe=promptPanel,
        boxrule=0pt,
        arc=2mm,
        left=2mm,
        right=2mm,
        top=1.5mm,
        bottom=1.5mm
    },
    promptpaneldark/.style={
        enhanced,
        colback=promptPanelDark,
        colframe=promptPanelDark,
        boxrule=0pt,
        arc=2mm,
        left=2mm,
        right=2mm,
        top=1.5mm,
        bottom=1.5mm
    }
}

\newtcblisting{promptcode}[1]{
    enhanced,
    listing only,
    title={\textbf{#1}},
    colback=promptCode,
    colframe=promptBorder,
    colbacktitle=promptPanelDark,
    coltitle=black,
    fonttitle=\bfseries\scriptsize,
    boxrule=0.6pt,
    arc=2mm,
    left=1mm,
    right=1mm,
    top=1mm,
    bottom=1mm,
    listing options={
        basicstyle=\ttfamily\fontsize{6.3pt}{7.1pt}\selectfont,
        columns=fullflexible,
        keepspaces=true,
        breaklines=true,
        breakatwhitespace=false,
        showstringspaces=false
    }
}

\usepackage{makecell}

\def\method{GTA-RAG}

\title{GTA-RAG: Graph-Trajectory-Augmented Reinforcement Learning for Multi-Turn Retrieval-Augmented Reasoning}

\author{
 \textbf{Jun Chen}$^{\textbf{1,2,3}}$ \hspace{4mm}
 \textbf{Yongchao Liu}$^{\textbf{3,}\boldsymbol{\dagger}}$ \hspace{4mm}
 \textbf{Pengyu Qiu}$^{\textbf{3}}$ \hspace{4mm}
 \textbf{Jiajun Zheng}$^{\textbf{3}}$ 
\\
 \textbf{Juelu Zhang}$^{\textbf{3}}$ \hspace{4mm}
 \textbf{Yujie Zeng}$^{\textbf{1,2,3}}$ \hspace{4mm}
 \textbf{Qin Zhang}$^{\textbf{1,}\boldsymbol{\dagger}}$ \hspace{4mm}
 \textbf{Ziyue Qiao}$^{\textbf{2,}\boldsymbol{\dagger}}$ \hspace{4mm}
 \textbf{Xiao Luo}$^{\textbf{4}}$ \hspace{4mm}
\vspace{4pt}
\\
 \textsuperscript{1}Shenzhen University, Shenzhen, China \\
 \textsuperscript{2}Great Bay University, Dongguan, China \\
 \textsuperscript{3}Ant Group, Hangzhou, China \\
 \textsuperscript{4}University of Wisconsin--Madison, Madison, USA
\\
 \small{
   $\boldsymbol{\dagger}$ Corresponding Authors: \href{mailto:zyqiao@gbu.edu.cn}{zyqiao@gbu.edu.cn}, \href{mailto:yongchao.ly@antgroup.com}{yongchao.ly@antgroup.com}, \href{mailto:qinzhang@szu.edu.cn}{qinzhang@szu.edu.cn}
 }
}

\begin{document}
\maketitle
\input{section0abstract_completed}
\input{section1introduction_completed}
\input{section2relatedwork_completed}
\input{section3method_graph_retrieval_revised}
\input{section4experiments_GTA_RAG_balanced}

\input{section5conclusion_completed}

\section*{Limitations}

Our framework relies on the quality of the constructed entity--document graph and the deployed retriever used for trajectory validation. Extraction noise or missing graph connections may limit the diversity of executable trajectories that can be generated. In addition, our experiments focus on open-domain QA benchmarks; extending \method{} to broader domains and more diverse retrieval environments remains an interesting direction for future work.

\section*{Acknowledgments}

The work of Ziyue Qiao is supported by the Guangdong Basic and Applied Basic Research Foundation (No. 2025A1515140238, No. 2024A1515140114), the National Natural Science Foundation of China (No. 62676075, No. 62406056), and Ant Group through CCF-Ant Research Fund. The work of Qin Zhang was partially supported by the National Natural Science Foundation of China (No. 62576221) and the Guangdong Provincial Natural Science Foundation (No. 2025A1515010288).


\bibliography{custom}
\input{appendix}




\end{document}

%% file: section0abstract_completed.tex
\begin{abstract}
Retrieval-augmented generation (RAG) enables LLMs to access external knowledge for answering knowledge-intensive questions. For complex multi-hop questions, multi-turn retrieval-augmented reasoning extends RAG into an iterative process that repeatedly searches for and integrates evidence across documents. However, existing reinforcement-learning (RL) approaches for agentic RAG are typically optimized with final-answer rewards, which provide sparse supervision and overlook whether the model actually retrieves the required evidence chain.
We present \textsc{GTA-RAG}, a graph-trajectory-augmented RL framework for multi-turn retrieval-augmented reasoning. From an entity--document graph, we sample connected document paths, synthesize multi-hop QA trajectories, and validate them with the deployed retriever to obtain executable trajectory-level supervision. We then optimize the retrieval policy with Group Relative Policy Optimization (GRPO) and a trajectory-guided reward that encourages both accurate answers and acquisition of target evidence documents, followed by answer-reward training on natural QA instances. Experiments on three multi-hop and two simple QA benchmarks show that \method{} consistently outperforms RL-based RAG baselines with both Qwen2.5-3B and Qwen2.5-7B backbones, while substantially improving evidence-chain coverage. Our code is available at \url{https://github.com/cjcj46262/GTA-RAG}.
\end{abstract}

%% file: section1introduction_completed.tex
\section{Introduction}
\label{sec:introduction}

Retrieval-augmented generation (RAG) equips large language models (LLMs) with external evidence and has become a practical paradigm for knowledge-intensive question answering~\citep{karpukhin2020denseRAG_2,lewis2020retrievalRAG_1,izacard2021leveragingRAG_3}. In multi-hop question answering, however, producing a correct response is not sufficient: a model must identify intermediate entities, issue successive searches, and integrate evidence distributed across multiple documents~\citep{chen2026pathrag,xu2025noderag}. A single retrieval step often fails because the document that directly contains the answer is only reachable after resolving one or more latent bridge entities~\citep{qian2025memorag}.

\begin{figure}[t]
    \centering
    \includegraphics[width=0.9\columnwidth]{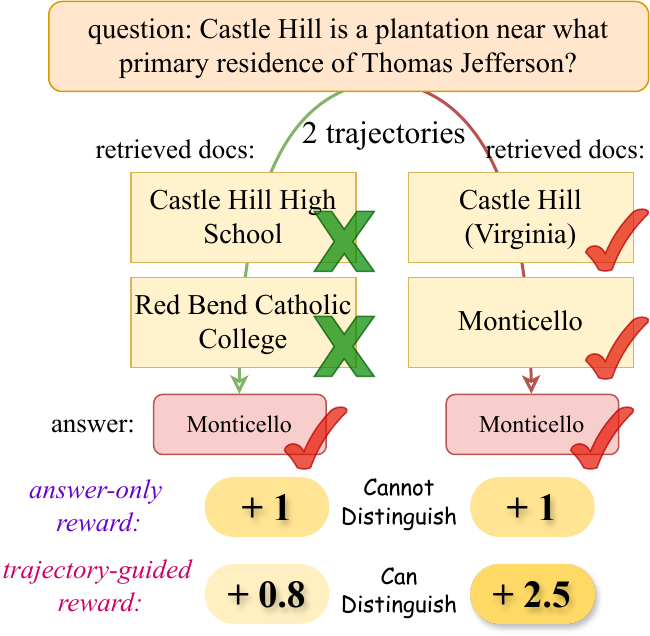}
    \caption{
    A correct answer can arise from an unsupported retrieval trajectory.
    Unlike answer-only reward, our trajectory-guided reward distinguishes it from an evidence-grounded trajectory.
    }
    \label{fig:shortcut_motivation}
    \vspace{-10pt}
\end{figure}

Recent RL-based agentic RAG methods have shown that LLMs can learn when and how to search through interaction with a retriever~\citep{yu2026graphragR1,endgraphGraphR1,tan2026rag_R1}. Nevertheless, most training signals are dominated by answer-level rewards, such as exact match (EM) on the final prediction. This objective is attractive because it is simple and broadly applicable, but it does not distinguish an evidence-grounded reasoning path from a shortcut that happens to output the correct answer. For example, in one observed case, a model correctly answers that Castle Hill is near \emph{Monticello} despite retrieving documents about an Australian high school and Australian locations rather than the required evidence documents, \emph{Castle Hill (Virginia)} and \emph{Monticello}. Rewarding this rollout solely by final-answer correctness reinforces an unfaithful retrieval behavior.

In this work, we argue that a graph retrieval environment provides a natural source of scalable process supervision. A graph knowledge base already exposes relations among entities and documents; connected document paths therefore encode potential evidence chains~\citep{gutierrez2024hipporag,xiang2025whentouse}. Rather than requiring manually annotated retrieval trajectories, we sample paths from the graph, prompt a strong generator to formulate multi-hop questions whose answers depend on the sampled evidence, and execute the generated search actions with the same retrieval interface used at training and inference time. Only trajectories whose intended intermediate documents are actually retrievable are retained. This \emph{trajectory validation} step converts graph connectivity into executable, retrieval-grounded training examples.

Based on these data, we propose \textbf{G}raph-\textbf{T}rajectory-\textbf{A}ugmented RAG (\method{}), an RL post-training framework for multi-turn retrieval-augmented reasoning. Our policy learns when and how to issue multi-turn search queries, while each retrieval call returns evidence obtained by combining graph-based retrieval with dense passage retrieval. We train the policy with GRPO~\citep{shao2024deepseekmathgrpo} in two stages. In the first stage, a \emph{trajectory-guided reward} gives intermediate feedback for retrieving new documents on the target evidence chain and provides an additional bonus when a correct answer is supported by a complete retrieved chain. In the second stage, the policy is further optimized with answer-level EM rewards on natural QA training instances, preserving the flexibility needed when explicit trajectories are unavailable.

Experiments on both multi-hop and simple QA benchmarks show that \method{} consistently outperforms prior RL-based RAG systems under both Qwen2.5-3B and Qwen2.5-7B backbones. Retrieval-behavior analysis further demonstrates that our policy more frequently recovers complete evidence chains with fewer unproductive search turns.

Our contributions are threefold:
\begin{itemize}
    \item We introduce a graph-based data augmentation procedure that automatically synthesizes and validates executable multi-hop retrieval trajectories from an entity--document graph.
    \item We propose a trajectory-guided GRPO objective that jointly encourages accurate answering and faithful evidence-chain acquisition in multi-turn RAG.
    \item We demonstrate consistent improvements on both multi-hop and simple QA benchmarks, and provide retrieval-level analyses showing that the gains are associated with better evidence coverage rather than answer-only shortcuts.
\end{itemize}


%% file: section2relatedwork_completed.tex
\section{Related Work}
\label{sec:related_work}

\subsection{Graph-Based Retrieval-Augmented Generation}
\label{sec:related_graph_rag}

Conventional RAG typically retrieves passages independently according to lexical or dense similarity with the input query~\citep{fan2024surveyRAG2,arslan2024surveyRAG1,ren2026context,peng2025graphsurvey}. This design is effective when the required evidence is locally similar to the question, but it is less suited to multi-hop queries in which the key evidence is connected through intermediate entities or relations. Graph-based RAG methods address this limitation by organizing corpus units into structured representations and retrieving through graph connectivity~\citep{zhang2025surveygraphRAG,guo2024lightrag,he2024Gretriever}. GraphRAG~\citep{edge2024graphRAG} constructs graph abstractions over a text collection to support structured retrieval and aggregation, while RAPTOR~\citep{sarthi2024raptor} recursively summarizes text into a retrieval hierarchy. HippoRAG~\citep{gutierrez2024hipporag} and its subsequent extensions use graph-style associations to improve multi-hop evidence discovery and memory-oriented retrieval.

Unlike methods that use graphs only for inference-time retrieval, we exploit graph paths to construct executable multi-document trajectories for supervising intermediate retrieval behavior.

\subsection{RL Post-Training for LLM Reasoning}
\label{sec:related_rl_rag}

Reinforcement learning has recently been used to improve the reasoning ability of LLMs through outcome-based optimization~\citep{endgraphGraphR1,yu2026graphragR1,song2025efficientRLgraphrag,lu2025scalingLLM}. Group Relative Policy Optimization (GRPO) provides an efficient policy-optimization formulation by estimating relative advantages from multiple sampled outputs for the same prompt, avoiding a separate value model. In retrieval-augmented settings, recent systems such as Search-R1, R1-Searcher, and RouteRAG train an LLM to interleave reasoning with search actions, demonstrating that RL can substantially improve information-seeking behavior.

However, answer-level rewards provide limited supervision for multi-turn RAG: a model may answer correctly from parametric knowledge or incomplete evidence without learning grounded retrieval behavior. Our trajectory-guided reward addresses this issue by providing intermediate feedback for acquiring documents along the target evidence chain constructed by our augmentation procedure.


%% file: section3method_graph_retrieval_revised.tex
\section{Method}
\label{sec:method}

\begin{figure*}[t]
    \centering
    \IfFileExists{figures/graph_augmented_RL_2140.pdf}{
        \includegraphics[width=0.999\textwidth]{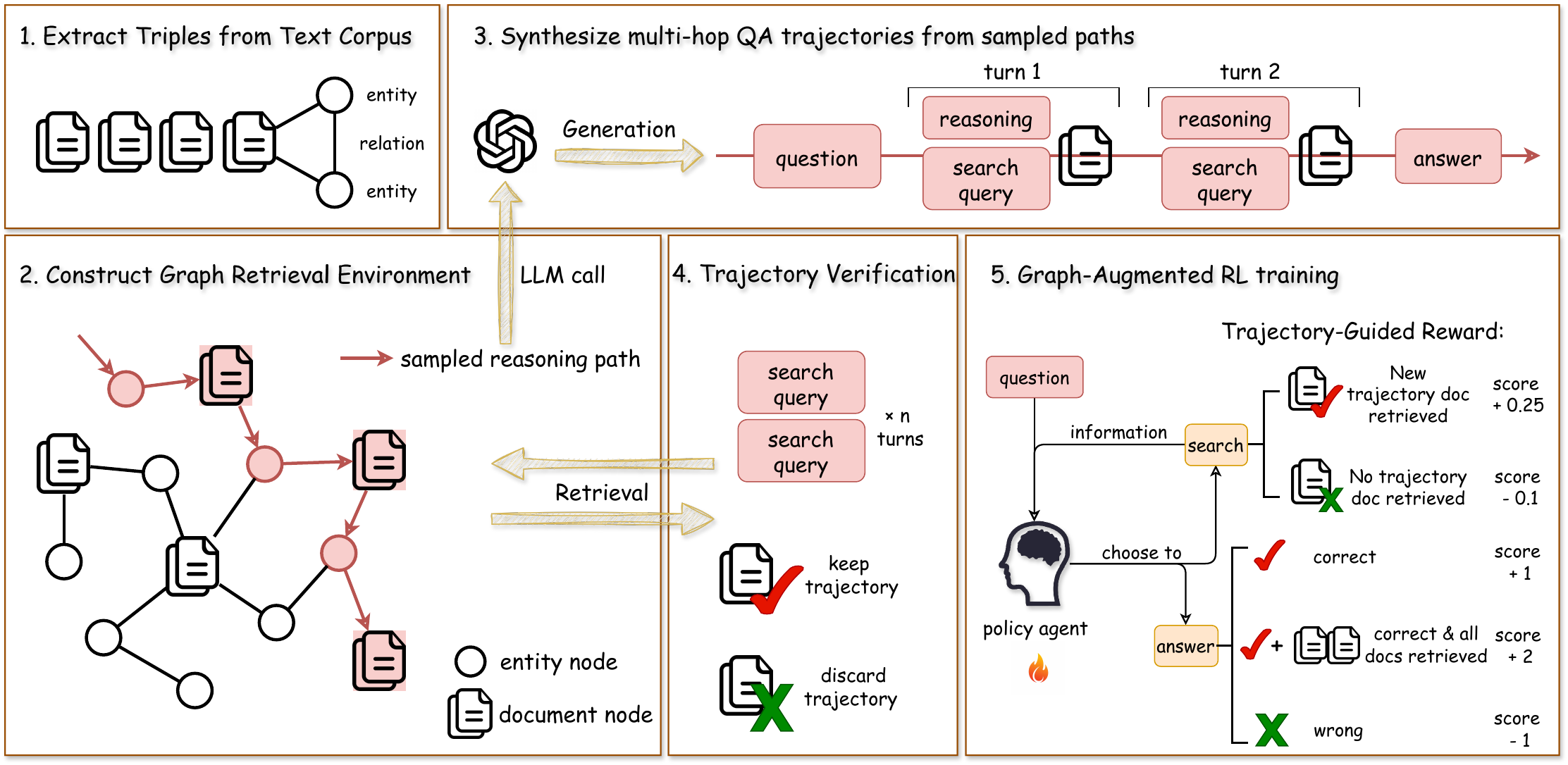}
    }{
        \fbox{\parbox[c][4.3cm][c]{0.96\textwidth}{
            \centering
            \textbf{Insert completed framework figure here: \texttt{figures/graph\_augmented\_RL\_2140.pdf}}\\[4pt]
            Panels: OpenIE triple extraction $\rightarrow$ entity--document graph construction $\rightarrow$ graph-path trajectory synthesis\\
            $\rightarrow$ retrieval validation $\rightarrow$ multi-turn GRPO training with trajectory-guided reward.
        }}
    }
    \vspace{-15pt}
    \caption{Overview of Graph-Trajectory-Augmented RAG. We first extract entity--relation--entity triples from the corpus and build a graph containing entity nodes, document nodes, relation edges, document--entity edges, and synonymy edges. Connected document paths sampled from this graph are converted into multi-hop question--answer trajectories by a generator model. Each synthetic trajectory is executed through the actual retrieval interface, which returns graph-retrieved evidence combined with dense passage results; trajectories that cannot retrieve their intended support documents are removed. Finally, an agent policy is optimized with GRPO, receiving intermediate trajectory-guided rewards for acquiring new target documents and outcome rewards for producing a correct answer supported by a complete evidence chain.}
    \label{fig:framework}
    \vspace{-10pt}
\end{figure*}

We propose \textbf{G}raph-\textbf{T}rajectory-\textbf{A}ugmented RAG, a framework that uses a graph-structured corpus both as a retrieval environment and as a source of trajectory-level supervision for reinforcement learning (RL) post-training. Figure~\ref{fig:framework} provides an overview. The framework consists of three components: (i) constructing an entity--document graph and a graph-augmented retriever; (ii) synthesizing and validating multi-hop retrieval trajectories from graph paths; and (iii) optimizing a multi-turn retrieval policy with a trajectory-guided GRPO objective followed by answer-reward adaptation.

\subsection{Problem Formulation}
\label{sec:problem_formulation}

Let $\mathcal{D}=\{d_i\}_{i=1}^{N}$ denotes a corpus of documents and $x$ a question. A retrieval-augmented policy $\pi_{\theta}$ generates a multi-turn trajectory
\begin{equation}
    \tau = (z_1, a_1, o_1, \ldots, z_T, a_T),
\end{equation}
where $z_t$ is the textual reasoning state available before action $a_t$, and $o_t$ is the observation returned by the retriever after a search action. At each turn, the policy either issues a retrieval query
\begin{equation}
    a_t =
    \texttt{\textless search\textgreater}\;
    q_t\;
    \texttt{\textless/search\textgreater},
\end{equation}
or terminates with a final prediction
\begin{equation}
    a_t =
    \texttt{\textless answer\textgreater}\;
    \hat{y}\;
    \texttt{\textless/answer\textgreater}.
\end{equation}
For a multi-hop retrieval question, let
$\mathcal{S}(x)=\{d^{\star}_1,\ldots,d^{\star}_H\}$
be its target evidence documents. Our objective is not only to maximize answer correctness, but also to encourage trajectories that retrieve the required evidence documents before answering.

\subsection{Graph Knowledge Base and Retrieval}
\label{sec:graph_kb}

\paragraph{Entity--document graph construction.}
Starting from each document $d_i\in\mathcal{D}$, we apply an OpenIE model to extract factual triples
\begin{equation}
    \mathcal{T}_i = \{(e_h, r, e_t)\},
\end{equation}
where $e_h$ and $e_t$ are head and tail entities and $r$ is a textual relation. We construct a heterogeneous graph
\begin{equation}
    \mathcal{G} = (\mathcal{V}_{D}\cup\mathcal{V}_{E},
    \mathcal{E}_{DE}\cup\mathcal{E}_{EE}^{r}\cup\mathcal{E}_{EE}^{s}),
\end{equation}
where $\mathcal{V}_{D}$ contains document nodes and $\mathcal{V}_{E}$ contains entity nodes. We add three types of edges: (1) a document--entity edge $(d_i,e)$ if entity $e$ is extracted from $d_i$; (2) a relation edge $(e_h,e_t)$ for each extracted triple; and (3) a synonymy edge between two entities whose embedding cosine similarity exceeds a threshold $\delta$. The synonymy edges reduce graph fragmentation caused by surface-form variations.

\paragraph{Graph retrieval.}
Given a search query $q$, we embed $q$ and all extracted triples and score each triple by cosine similarity:
\begin{equation}
    s(q,t)=\cos(\mathbf{h}_{q},\mathbf{h}_{t}),
    \qquad t\in\mathcal{T}.
\end{equation}
The top-scoring triples transfer their relevance scores to the participating entity nodes, yielding an initial personalization vector $\mathbf{p}_{q}$. Starting from these query-relevant entity nodes, we perform personalized PageRank (PPR) over the entity--document graph:
\begin{equation}
    \mathbf{r}_{q}
    =
    \alpha \mathbf{p}_{q}
    +
    (1-\alpha)\mathbf{P}^{\top}\mathbf{r}_{q},
    \label{eq:ppr}
\end{equation}
where $\mathbf{P}$ is the normalized transition matrix of $\mathcal{G}$ and $\alpha$ is the restart probability. The resulting scores propagate relevance through relational and entity--document connections, allowing the retriever to identify documents that may not be lexically similar to the current query but are structurally relevant to the evolving reasoning chain. We retain the top-$K_g$ document nodes as graph-retrieved evidence:
\begin{equation}
    \mathcal{R}_{g}(q)
    =
    \operatorname{TopK}_{d\in\mathcal{V}_{D}}
    \bigl(\mathbf{r}_{q}(d), K_g\bigr).
\end{equation}

\paragraph{Returned retrieval evidence.}
Although graph retrieval is the central retrieval mechanism in our framework, dense passage retrieval remains useful for recovering documents with strong semantic similarity to the current query. Therefore, for each query $q$, we additionally obtain the top-$K_d$ passage results $\mathcal{R}_{d}(q)$ from a dense retriever and return the deduplicated combination of both sources:
\begin{equation}
    \mathcal{R}(q)
    =
    \operatorname{Dedup}
    \left(
        \mathcal{R}_{g}(q)\cup\mathcal{R}_{d}(q)
    \right).
    \label{eq:combined_retrieval}
\end{equation}
Thus, the policy only needs to formulate a search query, while the retrieval interface consistently provides evidence enhanced by graph reasoning and dense passage matching.

\subsection{Multi-Turn Rollout Procedure}
\label{sec:multi_turn}

The policy operates as an interactive reasoning agent. At the beginning of a rollout, it observes only the input question. At each turn, the policy first generates a reasoning trace $z_t$ and then produces an action $a_t$, which is either a search query or a final answer. If a search action is issued, the retrieved documents are appended inside \texttt{\textless info\textgreater} tags, and the policy continues reasoning based only on the question, its previous reasoning traces, and the evidence retrieved so far. This procedure prevents leakage of future support documents during both synthetic trajectory construction and policy rollout.

Formally, the initial context is $c_1=x$. After $t-1$ search interactions, the context available to the policy is
\begin{equation}
    c_t =
    \left[
        x;
        z_1; a_1; o_1;
        \cdots;
        z_{t-1}; a_{t-1}; o_{t-1}
    \right].
\end{equation}
At turn $t$, the policy jointly generates a reasoning trace and an action:
\begin{equation}
    (z_t,a_t) \sim \pi_{\theta}(\cdot \mid c_t).
\end{equation}
If $a_t$ is a search action with query $q_t$, the retriever returns
$o_t=\mathcal{R}(q_t)$ and the context is updated as
\begin{equation}
    c_{t+1}
    =
    c_t
    \oplus z_t
    \oplus a_t
    \oplus
    \texttt{\textless info\textgreater}
    o_t
    \texttt{\textless/info\textgreater}.
\end{equation}

The rollout terminates when the policy emits an answer action or reaches the maximum action budget $B$.

\begin{algorithm}[t]
\caption{Multi-Turn Rollout Procedure}
\label{alg:rollout}
\footnotesize
\begin{algorithmic}[1]
\Require Question $x$, policy $\pi_{\theta}$, retriever $\mathcal{R}$, action budget $B$
\Ensure Final trajectory $\tau$ and predicted answer $\hat{y}$
\State $c \gets x$; $\tau \gets \varnothing$
\For{$t=1$ to $B$}
    \State Generate $(z_t,a_t) \sim \pi_{\theta}(\cdot \mid c)$ until 
    \texttt{\textless/search\textgreater} or 
    \texttt{\textless/answer\textgreater} is produced
    \If{$a_t$ matches \texttt{\textless search\textgreater} $q_t$ \texttt{\textless/search\textgreater}}
        \State $o_t \gets \mathcal{R}(q_t)$
        \State $c \gets c \oplus z_t \oplus a_t \oplus
        \texttt{\textless info\textgreater}
        o_t
        \texttt{\textless/info\textgreater}$
        \State $\tau \gets \tau \oplus (z_t,a_t,o_t)$
    \ElsIf{$a_t$ matches \texttt{\textless answer\textgreater} $\hat{y}$ \texttt{\textless/answer\textgreater}}
        \State $\tau \gets \tau \oplus (z_t,a_t)$
        \State \Return $\tau,\hat{y}$
    \Else
        \State $c \gets c \oplus z_t \oplus$ ``My action is invalid. Let me rethink.''
    \EndIf
\EndFor
\State \Return $\tau,\hat{y}=\varnothing$
\end{algorithmic}
\end{algorithm}

\subsection{Multi-hop QA Trajectory Construction}
\label{sec:trajectory_construction}

Optimizing a multi-turn retrieval policy with final-answer rewards alone provides sparse supervision: as shown in Figure~\ref{fig:shortcut_motivation}, a policy can answer correctly despite retrieving irrelevant documents, receiving the same reward as an evidence-grounded rollout. To provide intermediate retrieval supervision without manual annotation, we exploit the corpus graph to sample connected document paths, synthesize multi-hop QA trajectories with explicit target-document chains, and validate them using the deployed retriever.
\paragraph{Path sampling.}
To generate supervision without manual trajectory annotation, we sample document-connected paths from $\mathcal{G}$. Starting at a randomly selected document node, a random walk visits new nodes without replacement until it reaches $H$ document nodes, where $H$ controls the intended hop complexity. We restrict any two consecutive document nodes on a sampled path to be connected through at most two entity nodes. The resulting document sequence
\begin{equation}
    P=(d_1^{\star},d_2^{\star},\ldots,d_H^{\star})
\end{equation}
defines a candidate supporting-document chain.

\paragraph{Question and trajectory synthesis.}
For each sampled path, a generator LLM receives the documents on $P$ together with distractor documents and is instructed to produce a multi-hop question $x$, a gold answer $y^{\star}$, and a turn-by-turn reasoning trajectory whose search queries reveal documents sequentially. Importantly, the generated reasoning at turn $t$ is conditioned only on the question and documents revealed in earlier turns, matching the information constraints of policy rollouts.

\paragraph{Retriever-based validation.}
Graph connectivity alone does not guarantee that a generated search query can recover its intended target document. We therefore execute every synthetic search query with the same graph-augmented retrieval interface used during policy training and evaluation. A generated example is retained only when, at every search turn, its designated support document appears in the returned evidence. This filtering yields validated trajectory supervision
\begin{equation}
    \mathcal{D}_{\mathrm{traj}}
    =
    \{(x,y^{\star},\mathcal{S}(x),\tau^{\star})\},
\end{equation}
where $\tau^{\star}$ is executable by the deployed retriever.

\subsection{Trajectory-Guided GRPO}
\label{sec:graph_augmented_grpo}

The validated trajectories identify the target evidence chain for each synthetic question, enabling supervision beyond final-answer correctness. We therefore design a trajectory-guided reward that favors evidence-grounded rollouts over shortcut answers without the required supporting documents.

\paragraph{Two-stage RL post-training.}
We train the policy in two successive stages. Stage~I uses validated synthetic instances in $\mathcal{D}_{\mathrm{traj}}$ and trajectory-guided rewards, teaching the model to conduct evidence-grounded multi-turn retrieval. Stage~II uses natural QA instances with answer-level exact-match reward, improving task adaptation beyond synthetic graph paths while retaining the search behavior acquired in Stage~I.

\paragraph{Trajectory-guided reward.}
For a trajectory $\tau$, let $\mathcal{C}_{t}$ denote the set of target documents retrieved before turn $t$, and let
\begin{equation}
    \Delta_t
    =
    \bigl(
        \mathcal{R}(q_t)\cap \mathcal{S}(x)
    \bigr)
    \setminus \mathcal{C}_{t}
\end{equation}
be the newly acquired target documents at search turn $t$. We define the search reward as
\begin{equation}
    r^{\mathrm{search}}_t =
    \begin{cases}
        +0.25 \cdot |\Delta_t|, & |\Delta_t|>0,\\
        -0.10, & |\Delta_t|=0.
    \end{cases}
    \label{eq:search_reward}
\end{equation}
At the terminal answer turn, the answer reward is
\begin{equation}
    r^{\mathrm{ans}} =
    \begin{cases}
        -1, & \operatorname{EM}(\hat{y},y^{\star})=0,\\
        +1, & \operatorname{EM}(\hat{y},y^{\star})=1
              \land \mathcal{C}_{T}\not\supseteq\mathcal{S}(x),\\
        +2, & \operatorname{EM}(\hat{y},y^{\star})=1
              \land \mathcal{C}_{T}\supseteq\mathcal{S}(x).
    \end{cases}
    \label{eq:answer_reward}
\end{equation}
Thus, a correct answer obtains a larger terminal reward when the complete evidence chain has been retrieved, rather than receiving the same score as a shortcut answer. The total Stage~I reward is
\begin{equation}
    R_{\mathrm{traj}}(\tau)
    =
    \sum_{t:a_t\in\mathcal{A}_{\mathrm{search}}}
    r^{\mathrm{search}}_t
    +
    r^{\mathrm{ans}}.
\end{equation}
In Stage~II, we use the broader answer-only reward
\begin{equation}
    R_{\mathrm{EM}}(\tau)
    =
    \mathbb{I}
    \left[
        \operatorname{EM}(\hat{y},y^{\star})=1
    \right].
\end{equation}

\paragraph{GRPO objective.}
For each question $x$, GRPO~\citep{shao2024deepseekmathgrpo,guo2025deepseekr1} samples a group of $G$ trajectories $\{\tau_i\}_{i=1}^{G}$ and normalizes their rewards within the group:
\begin{equation}
    \hat{A}_i =
    \frac{
        R(\tau_i)-\operatorname{mean}_{j=1}^{G}R(\tau_j)
    }{
        \operatorname{std}_{j=1}^{G}R(\tau_j)+\epsilon
    }.
\end{equation}
The policy is optimized using the clipped surrogate objective with a KL regularizer to a reference policy $\pi_{\mathrm{ref}}$:
\begin{align}
    \mathcal{L}_{\mathrm{GRPO}}(\theta)
    &= -\mathbb{E}_{x,\{\tau_i\}}
    \Bigg[
    \frac{1}{G}\sum_{i=1}^{G}
    \min\Big(
        \rho_i(\theta)\hat{A}_i,
        \nonumber\\
        &\operatorname{clip}(\rho_i(\theta),1-\varepsilon,1+\varepsilon)\hat{A}_i
    \Big)
    \nonumber\\
    &-
    \beta D_{\mathrm{KL}}
    \left(
        \pi_{\theta}\Vert\pi_{\mathrm{ref}}
    \right)
    \Bigg],
    \label{eq:grpo}
\end{align}
where $\rho_i(\theta)$ is the likelihood ratio between the current and behavior policies over the sampled trajectory. In Stage~I, $R=R_{\mathrm{traj}}$; in Stage~II, $R=R_{\mathrm{EM}}$.


%% file: section4experiments_GTA_RAG_balanced.tex
\section{Experiments}
\label{sec:experiments}


\begin{table*}[t]
\centering
\caption{
Comparison of retrieval-augmented generation methods on multi-hop QA and simple QA benchmarks.
The best results within each backbone group are shown in \textbf{bold}.
}
\vspace{-5pt}
\label{tab:main_results}
\setlength{\tabcolsep}{4.7pt}
\renewcommand{\arraystretch}{1.12}
\small
\begin{adjustbox}{max width=\textwidth}
\begin{tabular}{l cc cc cc cc cc cc}
\toprule
\multirow{3}{*}{\textbf{Method}}
& \multicolumn{6}{c}{\textbf{Multi-hop QA}}
& \multicolumn{4}{c}{\textbf{Simple QA}}
& \multicolumn{2}{c}{\textbf{Average}} \\
\cmidrule(lr){2-7}\cmidrule(lr){8-11}\cmidrule(lr){12-13}
& \multicolumn{2}{c}{\textbf{HotpotQA}}
& \multicolumn{2}{c}{\textbf{2Wiki}}
& \multicolumn{2}{c}{\textbf{MuSiQue}}
& \multicolumn{2}{c}{\textbf{PopQA}}
& \multicolumn{2}{c}{\textbf{NQ}}
& \multicolumn{2}{c}{} \\
\cmidrule(lr){2-3}\cmidrule(lr){4-5}\cmidrule(lr){6-7}
\cmidrule(lr){8-9}\cmidrule(lr){10-11}\cmidrule(lr){12-13}
& \textbf{EM} & \textbf{F1} & \textbf{EM} & \textbf{F1}
& \textbf{EM} & \textbf{F1} & \textbf{EM} & \textbf{F1}
& \textbf{EM} & \textbf{F1} & \textbf{EM} & \textbf{F1} \\
\midrule
\rowcolor{modelbg}
\multicolumn{13}{c}{\textit{\textbf{GPT-4o-mini}}} \\
\midrule
Direct Inference & 28.6 & 41.0 & 30.2 & 36.3 & 11.2 & 22.0 & 16.1 & 22.7 & 35.2 & 52.7 & 24.3 & 34.9 \\
\multicolumn{13}{l}{\textit{\textbf{Training-free RAG}}} \\[-1pt]
GraphRAG & 51.4 & 67.6 & 45.7 & 61.0 & 27.0 & 42.0 & 30.7 & 51.3 & 38.0 & 55.5 & 38.6 & 55.5 \\
LightRAG & 9.9 & 20.2 & 2.5 & 12.1 & 2.0 & 9.3 & 1.9 & 14.8 & 2.8 & 15.4 & 3.8 & 14.4 \\
RAPTOR & 50.6 & 64.7 & 39.7 & 48.4 & 27.7 & 39.2 & 41.9 & 55.1 & 37.8 & 54.5 & 39.5 & 52.4 \\
HippoRAG & 46.3 & 60.0 & 59.4 & 67.3 & 24.0 & 35.9 & \textbf{42.5} & \textbf{56.2} & 37.2 & 52.5 & 41.9 & 54.4 \\
HippoRAG 2 & \textbf{56.3} & \textbf{71.1} & \textbf{60.5} & \textbf{69.7} & \textbf{35.0} & \textbf{49.3} & 41.7 & 55.7 & \textbf{43.4} & \textbf{60.0} & \textbf{47.4} & \textbf{61.2} \\
\midrule
\rowcolor{modelbg}
\multicolumn{13}{c}{\textit{\textbf{Qwen2.5-3B}}} \\
\midrule
Vanilla RAG & 29.5 & 41.8 & 19.7 & 27.4 & 10.3 & 17.5 & 30.3 & 41.6 & 18.1 & 31.8 & 21.6 & 32.0 \\
\multicolumn{13}{l}{\textit{\textbf{Training-free RAG}}} \\[-1pt]
HippoRAG 2 & 31.2 & 45.0 & 21.5 & 33.8 & 12.2 & 20.2 & 29.1 & 40.1 & 20.3 & 33.5 & 22.9 & 34.5 \\
Search-o1 & 18.7 & 26.3 & 16.9 & 20.9 & 3.9 & 10.5 & 17.1 & 23.8 & 19.9 & 29.1 & 15.3 & 22.1 \\
\multicolumn{13}{l}{\textit{\textbf{RL-based RAG}}} \\[-1pt]
Search-R1 & 45.2 & 56.9 & 42.4 & 50.8 & 22.2 & 30.9 & 45.8 & 53.3 & \textbf{46.2} & \textbf{54.8} & 40.4 & 49.3 \\
RouteRAG & 49.3 & 60.5 & 51.8 & 58.5 & 24.6 & 34.2 & \textbf{46.2} & 54.6 & 40.2 & 49.2 & 42.4 & 51.4 \\
\rowcolor{oursbg}
\textbf{\method{} (Ours)} & \textbf{52.9} & \textbf{63.9} & \textbf{57.3} & \textbf{64.1} & \textbf{29.0} & \textbf{38.2} & 42.7 & 51.7 & 40.6 & 50.3 & \textbf{44.5} & \textbf{53.6} \\
\midrule
\rowcolor{modelbg}
\multicolumn{13}{c}{\textit{\textbf{Qwen2.5-7B}}} \\
\midrule
Vanilla RAG & 29.1 & 47.5 & 17.2 & 33.3 & 11.4 & 23.0 & 26.3 & 37.8 & 11.0 & 30.6 & 19.0 & 34.4 \\
\multicolumn{13}{l}{\textit{\textbf{Training-free RAG}}} \\[-1pt]
HippoRAG 2 & 27.4 & 46.1 & 16.8 & 34.7 & 12.3 & 24.0 & 27.0 & 37.9 & 8.1 & 27.2 & 18.3 & 34.0 \\
Search-o1 & 13.5 & 19.1 & 6.4 & 7.9 & 2.9 & 7.7 & 4.7 & 7.2 & 18.1 & 27.5 & 9.1 & 13.9 \\
\multicolumn{13}{l}{\textit{\textbf{RL-based RAG}}} \\[-1pt]
R1-Searcher & 46.6 & 56.7 & 41.7 & 49.0 & 29.3 & 37.6 & 28.4 & 41.0 & 41.6 & 52.2 & 37.5 & 47.3 \\
Search-R1 & 51.0 & 62.0 & 51.8 & 58.9 & 32.0 & 40.8 & \textbf{51.3} & 57.1 & \textbf{56.8} & \textbf{65.3} & 48.6 & 56.8 \\
RouteRAG & 55.6 & 67.5 & 53.9 & 61.2 & 34.6 & 43.8 & 48.1 & 55.3 & 47.4 & 56.9 & 47.9 & 56.9 \\
\rowcolor{oursbg}
\textbf{\method{} (Ours)} & \textbf{57.2} & \textbf{69.1} & \textbf{55.4} & \textbf{62.9} & \textbf{36.5} & \textbf{46.2} & 50.4 & \textbf{57.2} & 52.9 & 61.5 & \textbf{50.5} & \textbf{59.3} \\
\bottomrule
\end{tabular}
\end{adjustbox}
\vspace{-7pt}
\end{table*}

\subsection{Experimental Setup}
\label{sec:experimental_setting}

\paragraph{Benchmarks and evaluation.}
Following the setting in~\citep{guo2025routerag,sorokin2025qQrag,luo2026hypergraphrag}, we evaluate on five widely adopted open-domain QA benchmarks. HotpotQA, 2WikiMultiHopQA (2Wiki), and MuSiQue require multi-hop reasoning over distributed evidence, while PopQA and Natural Questions (NQ) mainly evaluate direct factual retrieval. Following prior RAG evaluation protocols, we report exact match (EM) and token-level F1, together with their unweighted averages across datasets.

\paragraph{Baselines.}
For GPT-4o-mini, we report direct inference and representative training-free RAG systems, including GraphRAG~\citep{edge2024graphRAG}, LightRAG~\citep{guo2024lightrag}, RAPTOR~\citep{sarthi2024raptor}, HippoRAG~\citep{gutierrez2024hipporag}, and HippoRAG~2~\citep{gutierrez2025hipporag2}. For the trainable Qwen2.5-3B and Qwen2.5-7B policies~\citep{hui2024qwen25}, we compare with vanilla RAG, training-free RAG Search-o1~\citep{li-etal-2025-searcho1}, and RL-based retrieval agents such as Search-R1~\citep{jin2025searchR1}, RouteRAG~\citep{guo2025routerag}, and R1-Searcher~\citep{song2025R1searcher} where available.

\paragraph{Models and retrieval environment.}
We instantiate the policy with Qwen2.5-3B and Qwen2.5-7B. Our graph construction pipeline extracts OpenIE triples using Llama-3.1-8B. At each search step, PPR-based graph retrieval finds structurally connected documents, while Contriever passage retrieval provides semantically matched evidence; their deduplicated union is returned to the policy. Graph statistics are reported in Appendix~\ref{app:additional_statistics}.

\paragraph{Trajectory augmentation and training.}
We generate candidate multi-hop trajectories from sampled graph paths and retain 2,000 trajectories after retriever-based and rule-based validation. Stage~I performs 20 optimization steps with trajectory-guided GRPO to learn evidence-chain-oriented retrieval behavior. Stage~II performs another 40 steps on 10,000 HotpotQA instances using EM reward to adapt the policy to natural questions without explicit trajectories. Each step contains 128 prompts with five rollouts per prompt. Detailed trajectory statistics are provided in Appendix~\ref{app:additional_statistics}.

\subsection{Main Results}
\label{sec:main_results}

Table~\ref{tab:main_results} shows that \method{} consistently improves RL-trained RAG agents across policy scales. With Qwen2.5-3B and Qwen2.5-7B, it obtains average EM/F1 scores of 44.5/53.6 and 50.5/59.3, respectively, outperforming the strongest RL-based alternatives in both backbone groups. The advantage is therefore not tied to a particular model capacity.

More importantly, the improvements are concentrated on multi-hop QA. Averaged over the three multi-hop datasets, \method{} improves over RouteRAG by 4.5 EM and 4.3 F1 points with the 3B backbone, and by 1.7 EM and 1.9 F1 points with the 7B backbone. In contrast, performance on simple QA is competitive but not uniformly superior. This asymmetry matches the design motivation: direct factual questions often require only one semantically matched passage, whereas multi-hop questions depend on locating intermediate evidence before an answer-bearing document can be identified.

The GPT-4o-mini results serve as a reference for training-free RAG performance with a proprietary backbone rather than as a controlled comparison. Notably, after RL post-training, our method with Qwen2.5 becomes competitive with the strongest GPT-4o-mini-based training-free method, HippoRAG~2. This observation highlights that graph-derived trajectories and trajectory-guided rewards can substantially strengthen a trainable open-source policy, enabling it to approach the performance of RAG systems powered by a proprietary model. Within the controlled Qwen comparisons, the consistent multi-hop gains further demonstrate the effectiveness of our training framework. Section~\ref{sec:retrieval_behavior} examines whether these gains are accompanied by improved evidence-chain retrieval behavior.


\begin{table}[t]
\centering
\small
\caption{
Ablation results on Qwen2.5-7B. Multi-hop scores are averaged over HotpotQA, 2Wiki, and MuSiQue. Full-chain denotes complete target-evidence retrieval. 
}
\vspace{-5pt}
\label{tab:ablation}
\setlength{\tabcolsep}{4.0pt}
\renewcommand{\arraystretch}{1.08}
\begin{tabular}{lccc}
\toprule
\textbf{Variant}
& \makecell{\textbf{Multi-hop}\\\textbf{EM}}
& \makecell{\textbf{Multi-hop}\\\textbf{F1}}
& \makecell{\textbf{Full-chain}\\\textbf{(\%)}} \\
\midrule
\method{} & \textbf{49.7} & \textbf{59.4} & \textbf{74.1} \\
\textit{w/o TGR}   & 46.2 & 56.0 & 58.7 \\
\textit{w/o TV}    & 47.0 & 56.8 & 62.4 \\
\textit{w/o GR}    & 46.8 & 56.6 & 60.9 \\
\bottomrule
\end{tabular}
\vspace{-8pt}
\end{table}

\begin{table*}[t]
\centering
\small
\caption{
Retrieval behavior on 1,000 held-out trajectory-based multi-hop questions using Qwen2.5-7B.
Higher is better for coverage metrics; lower is better for zero-target turns and average search turns.
}
\vspace{-5pt}
\label{tab:retrieval_behavior}
\setlength{\tabcolsep}{7pt}
\renewcommand{\arraystretch}{1.10}
\begin{tabular}{l cccc}
\toprule
\textbf{Variant}
& \textbf{Target-doc} $\uparrow$
& \textbf{Complete-trajectory} $\uparrow$
& \textbf{Zero-target} $\downarrow$
& \textbf{Avg.\ search} $\downarrow$ \\
& \textbf{Coverage (\%)}
& \textbf{Coverage (\%)}
& \textbf{Turns}
& \textbf{Turns} \\
\midrule
w/o trajectory-guided reward
& 68.1 & 54.3 & 1,094 & 2.48 \\
w/o trajectory validation
& 72.4 & 61.8 & 978 & 2.46 \\
\rowcolor{oursbg}
\textbf{\method{}}
& \textbf{82.3} & \textbf{74.1} & \textbf{755} & \textbf{2.42} \\
\bottomrule
\end{tabular}
\vspace{-7pt}
\end{table*}

\subsection{Ablation Study}
\label{sec:ablation}

We ablate three components on Qwen2.5-7B: trajectory-guided reward, trajectory validation, and graph retrieval. In Table~\ref{tab:ablation}, \textit{TGR} denotes trajectory-guided reward, \textit{TV} denotes trajectory validation, and \textit{GR} denotes graph retrieval; the variant \textit{w/o GR} replaces graph retrieval with dense-only retrieval. All variants retain the same backbone and training scale, allowing the effect of each component to be examined under a controlled setting.

As shown in Table~\ref{tab:ablation}, removing TGR produces the largest degradation in full-chain coverage, decreasing it from 74.1\% to 58.7\%, together with a 3.4-point drop in multi-hop F1. This result supports our motivation that final-answer reward alone is insufficient: a model can obtain a correct answer without retrieving the required evidence chain, whereas trajectory-guided reward explicitly encourages progress toward grounded multi-hop retrieval.

Removing TV also leads to clear decreases in both answer quality and full-chain coverage. This indicates that graph-derived trajectories are useful only when their intended retrieval steps are executable under the deployed retriever; otherwise, noisy or unreachable target paths weaken the supervision signal. Finally, removing GR reduces multi-hop performance and evidence coverage, suggesting that graph-based retrieval helps identify later-hop documents that may be difficult to recover through semantic similarity alone. Overall, these results show that graph retrieval supplies structured evidence paths, trajectory validation ensures their reliability, and trajectory-guided reward effectively transfers them into retrieval behavior improvements.

\subsection{Retrieval Behavior Analysis}
\label{sec:retrieval_behavior}

The main results show that \method{} improves end-to-end QA performance, especially on multi-hop benchmarks. We further examine whether these improvements arise from the intended effect of trajectory-guided RL: encouraging the policy to retrieve documents along the target evidence chain rather than producing correct answers from incomplete or irrelevant contexts.

We conduct a controlled retrieval-behavior analysis on 1,000 held-out synthetic multi-hop questions with known target-document trajectories. All variants use the same policy backbone, retrieval environment, training instances, and GRPO configuration, differing only in the removed trajectory-related component. Specifically, \textit{w/o trajectory-guided reward} replaces our process-aware reward with answer correctness alone, while \textit{w/o trajectory validation} trains on synthesized graph-path examples without filtering whether their target documents can be recovered by the deployed retriever.

We report four retrieval-oriented metrics. Let $\mathcal{Q}$ denote the evaluation set, $\mathcal{S}(x)$ the target document set for question $x$, and $\mathcal{C}(x)$ the set of documents retrieved by the policy before answering. \textbf{Target-document coverage} measures the proportion of target evidence documents retrieved across all examples:
\begin{equation}
    \mathrm{TargetDocCov}
    =
    \frac{
        \sum_{x \in \mathcal{Q}}
        \left|
            \mathcal{C}(x) \cap \mathcal{S}(x)
        \right|
    }{
        \sum_{x \in \mathcal{Q}}
        \left|
            \mathcal{S}(x)
        \right|
    }.
    \label{eq:target_doc_coverage}
\end{equation}
\textbf{Complete-trajectory coverage} measures the percentage of examples for which all target documents are retrieved before the final answer:
\begin{equation}
    \mathrm{CompleteTrajCov}
    =
    \frac{1}{|\mathcal{Q}|}
    \sum_{x \in \mathcal{Q}}
    \mathbb{I}
    \left[
        \mathcal{S}(x) \subseteq \mathcal{C}(x)
    \right].
    \label{eq:complete_trajectory_coverage}
\end{equation}
We additionally report \textbf{zero-target turns}, i.e., search turns that retrieve no previously unseen target document, and \textbf{average search turns}, which helps determine whether higher coverage is achieved through more effective retrieval rather than simply more retrieval calls.

Table~\ref{tab:retrieval_behavior} shows that trajectory-guided RL substantially improves evidence-chain acquisition. Compared with answer-only RL, \method{} increases target-document coverage from 68.1\% to 82.3\% and complete-trajectory coverage from 54.3\% to 74.1\%, indicating that it more frequently recovers the entire evidence chain rather than isolated relevant documents.

Removing trajectory validation decreases both coverage metrics, showing that graph-derived trajectories are effective only when executable by the deployed retriever. Moreover, \method{} reduces zero-target turns while using slightly fewer searches, demonstrating that its gains arise from more targeted and evidence-grounded retrieval rather than additional retrieval calls.


%% file: section5conclusion_completed.tex
\section{Conclusion}
\label{sec:conclusion}

We presented \method{}, an RL post-training framework that leverages graph-sampled and validated retrieval trajectories to improve multi-turn retrieval-augmented reasoning. By combining graph-augmented retrieval with trajectory-guided GRPO, our method encourages the policy to retrieve evidence along the required reasoning chain rather than relying on answer-only shortcuts. Experiments on multi-hop and simple QA benchmarks demonstrate consistent improvements over strong RL-based RAG baselines, while retrieval-behavior analysis further shows higher target-document and complete-trajectory coverage. These results highlight the potential of graph structure as scalable supervision for training more effective and evidence-grounded RAG agents.

%% file: appendix.tex
\vspace{0pt}
\newpage
\appendix

\section{Additional Experimental Details}
\label{app:additional_statistics}

\paragraph{Validated trajectory statistics.}
Table~\ref{tab:trajectory_statistics} summarizes the validated synthetic trajectories used for trajectory-guided RL training.

\paragraph{Graph statistics.}
Table~\ref{tab:graph_statistics} reports the scale of the entity--document graphs constructed for each benchmark corpus. In addition to the evaluation corpora, we construct a larger HotpotQA-10K graph for trajectory augmentation during training.


\begin{table}[h]
\vspace{-2pt}
\centering
\small
\caption{Statistics of the validated synthetic trajectory training set.}
\vspace{-8pt}
\label{tab:trajectory_statistics}
\begin{tabular}{l r}
\toprule
Statistic & Value \\
\midrule
Validated trajectories & 2,000 \\
2-hop trajectories & 1,899 \\
3-hop trajectories & 93 \\
4-hop trajectories & 8 \\
Maximum sampled document nodes per path & 10 \\
Stage I steps / prompts per step & 20 / 128 \\
Stage II steps / prompts per step & 40 / 128 \\
GRPO rollouts per prompt & 5 \\
\bottomrule
\end{tabular}
\end{table}

\begin{table*}[h]
\centering
\small
\caption{Statistics of the entity--document graphs used for graph retrieval and trajectory construction.}
\label{tab:graph_statistics}
\setlength{\tabcolsep}{7pt}
\renewcommand{\arraystretch}{1.08}
\begin{tabular}{lrrrrrr}
\toprule
\textbf{Dataset}
& \textbf{Document}
& \textbf{Entity}
& \textbf{Document--entity}
& \textbf{Relation}
& \textbf{Synonymy}
& \textbf{Avg.\ doc.} \\
& \textbf{nodes}
& \textbf{nodes}
& \textbf{edges}
& \textbf{edges}
& \textbf{edges}
& \textbf{degree} \\
\midrule
2WikiMultiHopQA & 6,119  & 44,356  & 61,219  & 112,252 & 207,288 & 10.00 \\
HotpotQA        & 9,811  & 84,890  & 122,182 & 227,492 & 199,187 & 12.45 \\
HotpotQA-10K    & 18,300 & 127,542 & 206,473 & 382,750 & 748,429 & 11.28 \\
MuSiQue         & 11,656 & 88,841  & 131,962 & 240,462 & 170,440 & 11.32 \\
NQ-REAR         & 9,633  & 76,971  & 132,168 & 224,324 & 159,393 & 13.72 \\
PopQA           & 8,676  & 75,906  & 114,018 & 199,618 & 152,737 & 13.14 \\
\bottomrule
\end{tabular}
\end{table*}

\section{Prompt Templates}
\label{app:prompt_templates}

\begin{table*}[p]
\centering
\caption{Prompt template for synthesizing validated multi-hop retrieval trajectories from graph-sampled document paths. Runtime-populated fields are highlighted in blue.}
\label{tab:trajectory_generation_prompt}
\begin{tcolorbox}[
    promptouter,
    title={Prompt Template for Graph-Augmented Trajectory Construction}
]
\scriptsize

\begin{tcolorbox}[promptpanel]
\textbf{Task Instruction.}
You are constructing high-quality multi-hop retrieval QA data. You are given a sampled graph path and a list of candidate documents from that path. Your job is to decide whether these documents can support a natural 2-hop, 3-hop, or 4-hop QA example.
\end{tcolorbox}

\vspace{1.5mm}

\begin{tcolorbox}[promptpaneldark]
\textbf{Retrieval Behavior.}
Dense passage retrieval finds documents by semantic similarity between the query and document text. Graph retrieval finds documents through ranked facts, linked entities, and graph reasoning over the structured knowledge graph.
\end{tcolorbox}

\vspace{1.2mm}

\begin{minipage}[t]{0.41\textwidth}
\vspace{0pt}

\begin{tcolorbox}[promptpanel]
\textbf{Construction Rules.}
\vspace{0.3mm}
\begin{enumerate}[leftmargin=4mm,itemsep=0.35mm,topsep=0.4mm]
    \item Use only the provided candidate documents; do not use external knowledge.
    \item Construct a natural bridge, comparison, or entity-linking question.
    \item The answer must be a short span supported by the selected documents.
    \item Use \promptslot{min\_hops} to \promptslot{max\_hops} search rounds.
    \item Each round must target one unique candidate document.
    \item The first query must be justified by the question alone.
    \item Later queries must be justified by previously retrieved evidence.
    \item Do not use unnecessary documents.
    \item Output \texttt{constructible=false} if no valid sample can be built.
    \item Do not reveal the answer in the question or search queries.
\end{enumerate}

\vspace{0.8mm}
\textbf{Action Format.}

\vspace{0.8mm}
\centering
\prompttag{\textless search\textgreater}
\;\texttt{query}\;
\prompttag{\textless/search\textgreater}

\vspace{1mm}

\prompttag{\textless answer\textgreater}
\;\texttt{answer}\;
\prompttag{\textless/answer\textgreater}
\end{tcolorbox}

\vspace{1.2mm}

\begin{promptcode}{Runtime Inputs}
Sampled path:
{sampled_path}

Candidate documents:
{candidate_documents}
\end{promptcode}

\end{minipage}
\hfill
\begin{minipage}[t]{0.57\textwidth}
\vspace{0pt}

\begin{promptcode}{Required JSON Output Schema}
{
  "constructible": true or false,
  "reason_if_not_constructible": "...",
  "question": "...",
  "gold_answer": ["..."],
  "hop_count": 2 or 3 or 4,
  "reasoning_process": [
    {
      "round_id": 0,
      "visible_information": "question only",
      "reasoning": "...",
      "action_type": "search",
      "search_query": "...",
      "action": "<search> ... </search>"
    },
    {
      "round_id": 1,
      "visible_information":
          "question + previous support documents",
      "reasoning": "...",
      "action_type": "search",
      "search_query": "...",
      "action": "<search> ... </search>"
    },
    {
      "round_id": 2,
      "visible_information":
          "question + all support documents",
      "reasoning": "...",
      "action_type": "answer",
      "action": "<answer> ... </answer>"
    }
  ],
  "support_docs": [
    {
      "turn_id": 0,
      "search_query": "...",
      "expected_doc_id": "d0",
      "expected_doc_title": "..."
    }
  ]
}
\end{promptcode}

\end{minipage}




\end{tcolorbox}
\end{table*}

%% file: custom.bib
@inproceedings{ren2026context,
  title={When Context Bites: Detecting RAG Poisoning via Document-Level Attention Collapse},
  author={Ren, Yingtao and Zhao, Ziyi and Fu, Yiwei and Luo, Xiao and Chang, Yu-Cheng and Lin, Chin-Teng},
  booktitle={Proceedings of the 49th International ACM SIGIR Conference on Research and Development in Information Retrieval},
  pages={4080--4086},
  year={2026}
}

@article{peng2025graphsurvey,
  title={Graph retrieval-augmented generation: A survey},
  author={Peng, Boci and Zhu, Yun and Liu, Yongchao and Bo, Xiaohe and Shi, Haizhou and Hong, Chuntao and Zhang, Yan and Tang, Siliang},
  journal={ACM Transactions on Information Systems},
  volume={44},
  number={2},
  pages={1--52},
  year={2025},
  publisher={ACM New York, NY}
}

@article{zhang2025surveygraphRAG,
  title={A survey of graph retrieval-augmented generation for customized large language models},
  author={Zhang, Qinggang and Chen, Shengyuan and Bei, Yuanchen and Yuan, Zheng and Zhou, Huachi and Hong, Zijin and Chen, Hao and Xiao, Yilin and Zhou, Chuang and Dong, Junnan and others},
  journal={arXiv preprint arXiv:2501.13958},
  year={2025}
}

@article{arslan2024surveyRAG1,
  title={A Survey on RAG with LLMs},
  author={Arslan, Muhammad and Ghanem, Hussam and Munawar, Saba and Cruz, Christophe},
  journal={Procedia computer science},
  volume={246},
  pages={3781--3790},
  year={2024},
  publisher={Elsevier}
}

@inproceedings{fan2024surveyRAG2,
  title={A survey on rag meeting llms: Towards retrieval-augmented large language models},
  author={Fan, Wenqi and Ding, Yujuan and Ning, Liangbo and Wang, Shijie and Li, Hengyun and Yin, Dawei and Chua, Tat-Seng and Li, Qing},
  booktitle={Proceedings of the 30th ACM SIGKDD conference on knowledge discovery and data mining},
  pages={6491--6501},
  year={2024}
}

@article{edge2024graphRAG,
  title={From local to global: A graph rag approach to query-focused summarization},
  author={Edge, Darren and Trinh, Ha and Cheng, Newman and Bradley, Joshua and Chao, Alex and Mody, Apurva and Truitt, Steven and Metropolitansky, Dasha and Ness, Robert Osazuwa and Larson, Jonathan},
  journal={arXiv preprint arXiv:2404.16130},
  year={2024}
}

@article{gutierrez2024hipporag,
  title={Hipporag: Neurobiologically inspired long-term memory for large language models},
  author={Guti{\'e}rrez, Bernal J and Shu, Yiheng and Gu, Yu and Yasunaga, Michihiro and Su, Yu},
  journal={Advances in neural information processing systems},
  volume={37},
  pages={59532--59569},
  year={2024}
}

@inproceedings{gutierrez2025hipporag2,
  title     = {From {RAG} to Memory: Non-Parametric Continual Learning for Large Language Models},
  author    = {Guti{\'e}rrez, Bernal Jim{\'e}nez and Shu, Yiheng and Qi, Weijian and Zhou, Sizhe and Su, Yu},
  booktitle = {Proceedings of the 42nd International Conference on Machine Learning},
  pages     = {21497--21515},
  year      = {2025},
  volume    = {267},
  series    = {Proceedings of Machine Learning Research},
  publisher = {PMLR}
}

@inproceedings{li-etal-2025-searcho1,
  title     = {Search-o1: Agentic Search-Enhanced Large Reasoning Models},
  author    = {Li, Xiaoxi and Dong, Guanting and Jin, Jiajie and Zhang, Yuyao and Zhou, Yujia and Zhu, Yutao and Zhang, Peitian and Dou, Zhicheng},
  booktitle = {Proceedings of the 2025 Conference on Empirical Methods in Natural Language Processing},
  pages     = {5420--5438},
  year      = {2025},
  doi       = {10.18653/v1/2025.emnlp-main.276},
  url       = {https://aclanthology.org/2025.emnlp-main.276/}
}

@article{he2024Gretriever,
  title={G-retriever: Retrieval-augmented generation for textual graph understanding and question answering},
  author={He, Xiaoxin and Tian, Yijun and Sun, Yifei and Chawla, Nitesh V and Laurent, Thomas and LeCun, Yann and Bresson, Xavier and Hooi, Bryan},
  journal={Advances in Neural Information Processing Systems},
  volume={37},
  pages={132876--132907},
  year={2024}
}

@inproceedings{sarthi2024raptor,
  title={Raptor: Recursive abstractive processing for tree-organized retrieval},
  author={Sarthi, Parth and Abdullah, Salman and Tuli, Aditi and Khanna, Shubh and Goldie, Anna and Manning, Christopher},
  booktitle={International Conference on Learning Representations},
  volume={2024},
  pages={32628--32649},
  year={2024}
}

@article{luo2026hypergraphrag,
  title={Hypergraphrag: Retrieval-augmented generation via hypergraph-structured knowledge representation},
  author={Luo, Haoran and Chen, Guanting and Zheng, Yandan and Wu, Xiaobao and Guo, Yikai and Lin, Qika and Feng, Yu and Kuang, Zemin and Song, Meina and Zhu, Yifan and others},
  journal={Advances in Neural Information Processing Systems},
  volume={38},
  pages={152206--152234},
  year={2025}
}

@article{guo2024lightrag,
  title={Lightrag: Simple and fast retrieval-augmented generation},
  author={Guo, Zirui and Xia, Lianghao and Yu, Yanhua and Ao, Tian and Huang, Chao},
  journal={arXiv preprint arXiv:2410.05779},
  volume={2},
  number={3},
  year={2024}
}

@inproceedings{qian2025memorag,
  title={Memorag: Boosting long context processing with global memory-enhanced retrieval augmentation},
  author={Qian, Hongjin and Liu, Zheng and Zhang, Peitian and Mao, Kelong and Lian, Defu and Dou, Zhicheng and Huang, Tiejun},
  booktitle={Proceedings of the ACM on Web Conference 2025},
  pages={2366--2377},
  year={2025}
}

@article{xu2025noderag,
  title={NodeRAG: Structuring graph-based rag with heterogeneous nodes},
  author={Xu, Tianyang and Zheng, Haojie and Li, Chengze and Chen, Haoxiang and Liu, Yixin and Chen, Ruoxi and Sun, Lichao},
  journal={arXiv preprint arXiv:2504.11544},
  year={2025}
}

@inproceedings{chen2026pathrag,
  title={Pathrag: Pruning graph-based retrieval augmented generation with relational paths},
  author={Chen, Boyu and Guo, Zirui and Yang, Zidan and Chen, Yuluo and Chen, Junze and Liu, Zhenghao and Shi, Chuan and Yang, Cheng},
  booktitle={Proceedings of the AAAI conference on artificial intelligence},
  volume={40},
  number={36},
  pages={30183--30191},
  year={2026}
}

@inproceedings{tan2026rag_R1,
  title={Rag-r1: Incentivizing the search and reasoning capabilities of llms through multi-query parallelism},
  author={Tan, Zhiwen and Huang, Jiaming and Wu, Qintong and Zhang, Hongxuan and Zhuang, Chenyi and Gu, Jinjie},
  booktitle={Proceedings of the AAAI Conference on Artificial Intelligence},
  volume={40},
  number={39},
  pages={33187--33195},
  year={2026}
}

@article{song2025efficientRLgraphrag,
  title={Efficient and transferable agentic knowledge graph rag via reinforcement learning},
  author={Song, Jinyeop and Wang, Song and Shun, Julian and Zhu, Yada},
  journal={arXiv preprint arXiv:2509.26383},
  year={2025}
}

@inproceedings{yu2026graphragR1,
  title={Graphrag-r1: Graph retrieval-augmented generation with process-constrained reinforcement learning},
  author={Yu, Chuanyue and Zhao, Kuo and Li, Yuhan and Chang, Heng and Feng, Mingjian and Jiang, Xiangzhe and Sun, Yufei and Li, Jia and Zhang, Yuzhi and Sun, Qingyun and others},
  booktitle={Proceedings of the ACM Web Conference 2026},
  pages={1398--1409},
  year={2026}
}

@article{endgraphGraphR1,
  title={Graph-R1: Towards Agentic GraphRAG Framework via End-to-end Reinforcement Learning},
  author={Luo, Haoran and E, Haihong and Chen, Guanting and Lin, Qika and Guo, Yikai and Xu, Fangzhi and Kuang, Zemin and Song, Meina and Wu, Xiaobao and Zhu, Yifan and Tuan, Luu Anh},
  journal={arXiv preprint arXiv:2507.21892},
  year={2025}
}

@article{jin2025searchR1,
  title={Search-r1: Training llms to reason and leverage search engines with reinforcement learning},
  author={Jin, Bowen and Zeng, Hansi and Yue, Zhenrui and Yoon, Jinsung and Arik, Sercan and Wang, Dong and Zamani, Hamed and Han, Jiawei},
  journal={arXiv preprint arXiv:2503.09516},
  year={2025}
}

@article{guo2025routerag,
  title={RouteRAG: Efficient Retrieval-Augmented Generation from Text and Graph via Reinforcement Learning},
  author={Guo, Yucan and Su, Miao and Guan, Saiping and Sun, Zihao and Jin, Xiaolong and Guo, Jiafeng and Cheng, Xueqi},
  journal={arXiv preprint arXiv:2512.09487},
  year={2025}
}

@article{sorokin2025qQrag,
  title={Q-RAG: Long Context Multi-step Retrieval via Value-based Embedder Training},
  author={Sorokin, Artyom and Buzun, Nazar and Anokhin, Alexander and Inozemcev, Oleg and Vedernikov, Egor and Anokhin, Petr and Burtsev, Mikhail and Alexey, Trushkov and Wenshuai, Yin and Burnaev, Evgeny},
  journal={arXiv preprint arXiv:2511.07328},
  year={2025}
}

@article{lu2025scalingLLM,
  title={Scaling llm multi-turn rl with end-to-end summarization-based context management},
  author={Lu, Miao and Sun, Weiwei and Du, Weihua and Ling, Zhan and Yao, Xuesong and Liu, Kang and Chen, Jiecao},
  journal={arXiv preprint arXiv:2510.06727},
  year={2025}
}

@article{song2025R1searcher,
  title={R1-searcher: Incentivizing the search capability in llms via reinforcement learning},
  author={Song, Huatong and Jiang, Jinhao and Min, Yingqian and Chen, Jie and Chen, Zhipeng and Zhao, Wayne Xin and Fang, Lei and Wen, Ji-Rong},
  journal={arXiv preprint arXiv:2503.05592},
  year={2025}
}

@article{shao2024deepseekmathgrpo,
  title={Deepseekmath: Pushing the limits of mathematical reasoning in open language models},
  author={Shao, Zhihong and Wang, Peiyi and Zhu, Qihao and Xu, Runxin and Song, Junxiao and Bi, Xiao and Zhang, Haowei and Zhang, Mingchuan and Li, YK and Wu, Yang and others},
  journal={arXiv preprint arXiv:2402.03300},
  year={2024}
}

@article{guo2025deepseekr1,
  title={Deepseek-r1: Incentivizing reasoning capability in llms via reinforcement learning},
  author={Guo, Daya and Yang, Dejian and Zhang, Haowei and Song, Junxiao and Wang, Peiyi and Zhu, Qihao and Xu, Runxin and Zhang, Ruoyu and Ma, Shirong and Bi, Xiao and others},
  journal={arXiv preprint arXiv:2501.12948},
  year={2025}
}

@article{hui2024qwen25,
  title={Qwen2. 5-coder technical report},
  author={Hui, Binyuan and Yang, Jian and Cui, Zeyu and Yang, Jiaxi and Liu, Dayiheng and Zhang, Lei and Liu, Tianyu and Zhang, Jiajun and Yu, Bowen and Lu, Keming and others},
  journal={arXiv preprint arXiv:2409.12186},
  year={2024}
}

@article{lewis2020retrievalRAG_1,
  title={Retrieval-augmented generation for knowledge-intensive nlp tasks},
  author={Lewis, Patrick and Perez, Ethan and Piktus, Aleksandra and Petroni, Fabio and Karpukhin, Vladimir and Goyal, Naman and K{\"u}ttler, Heinrich and Lewis, Mike and Yih, Wen-tau and Rockt{\"a}schel, Tim and others},
  journal={Advances in neural information processing systems},
  volume={33},
  pages={9459--9474},
  year={2020}
}

@inproceedings{karpukhin2020denseRAG_2,
  title={Dense passage retrieval for open-domain question answering},
  author={Karpukhin, Vladimir and Oguz, Barlas and Min, Sewon and Lewis, Patrick and Wu, Ledell and Edunov, Sergey and Chen, Danqi and Yih, Wen-tau},
  booktitle={Proceedings of the 2020 conference on empirical methods in natural language processing (EMNLP)},
  pages={6769--6781},
  year={2020}
}

@inproceedings{izacard2021leveragingRAG_3,
  title={Leveraging passage retrieval with generative models for open domain question answering},
  author={Izacard, Gautier and Grave, Edouard},
  booktitle={Proceedings of the 16th conference of the european chapter of the association for computational linguistics: main volume},
  pages={874--880},
  year={2021}
}

@article{xiang2025whentouse,
  title={When to use graphs in rag: A comprehensive analysis for graph retrieval-augmented generation},
  author={Xiang, Zhishang and Wu, Chuanjie and Zhang, Qinggang and Chen, Shengyuan and Hong, Zijin and Huang, Xiao and Su, Jinsong},
  journal={arXiv preprint arXiv:2506.05690},
  year={2025}
}
